\pdfoutput=1

\documentclass[11pt]{article}

\usepackage[preprint]{acl}

\usepackage{times}
\usepackage{latexsym}

\usepackage[T1]{fontenc}

\usepackage[utf8]{inputenc}

\usepackage{microtype}

\usepackage{inconsolata}

\usepackage{graphicx}
\usepackage{amsmath}
\usepackage{amsfonts}
\usepackage{amssymb}
\usepackage{cleveref}
\usepackage{colortbl}
\usepackage{booktabs}
\usepackage{multirow}
\usepackage{soul}
\usepackage{tabularx}
\usepackage{xspace}
\usepackage{microtype}
\usepackage{pifont}
\usepackage{subcaption}  %
\usepackage{float}
\usepackage{placeins}

\newcommand{\MODEL}{\mbox{\textsc{HDPO}}\xspace}
\newcommand{\A}{\mbox{\textsc{VDH}}\xspace}
\newcommand{\B}{\mbox{\textsc{LCH}}\xspace}
\newcommand{\C}{\mbox{\textsc{MCH}}\xspace}
\usepackage{xcolor}

\title{Mitigating Hallucination in Multimodal Large Language Model via Hallucination-targeted Direct Preference Optimization}

\author{
    Yuhan Fu$^{1,2}$\thanks{Equal contribution.},\quad
    Ruobing Xie$^{2}$\footnotemark[1],\quad
    Xingwu Sun$^{2}$,\quad
    Zhanhui Kang$^{2}$,\quad
    Xirong Li$^{1}$\thanks{Corresponding authors.} \\
    $^1$Key Lab of DEKE, Renmin University of China \\
    $^2$Machine Learning Platform Department, Tencent \\
    \texttt{\{fuyuhan, xirong\}@ruc.edu.cn} \quad  \texttt{ruobingxie@tencent.com}
}

\begin{document}
\maketitle

\begin{abstract}
Multimodal Large Language Models (MLLMs) are known to hallucinate, which limits their practical applications.
Recent works have attempted to apply Direct Preference Optimization (DPO) to enhance the performance of MLLMs, but have shown inconsistent improvements in mitigating hallucinations.
To address this issue more effectively, we introduce Hallucination-targeted Direct Preference Optimization (\MODEL) to reduce hallucinations in MLLMs.
Unlike previous approaches, our method tackles hallucinations from their diverse forms and causes.
Specifically, we develop three types of preference pair data targeting the following causes of MLLM hallucinations: (1) insufficient visual capabilities, (2) long context generation, and (3) multimodal conflicts.
Experimental results demonstrate that our method achieves superior performance across multiple hallucination evaluation datasets, surpassing most state-of-the-art (SOTA) methods and highlighting the potential of our approach.
Ablation studies and in-depth analyses further confirm the effectiveness of our method and suggest the potential for further improvements through scaling up.
\end{abstract}

\section{Introduction}

Large Language Models (LLMs) have been verified in various field \cite{OpenAI2024GPT4o,dubey2024llama,sun2024hunyuan}, while they encounter challenges such as hallucination.
Multimodal Large Language Models (MLLMs) are also known to hallucinate \cite{bai2024hallucination}.
Specifically, they often produce unfaithful content that does not align with the visual input, 
which undermines their reliability and practicality, particularly in critical applications such as autonomous driving \cite{cui2024survey} or medical tasks \cite{liu2023medical}.
Hence, addressing MLLM hallucination (\textbf{M-hallu}) is essential.

Recent efforts have aimed at mitigating M-hallu through various approaches, including inference-stage strategies like contrastive decoding \cite{leng2024mitigating}, and post-hoc corrections that employ external visual models to refine responses \cite{yin2023woodpecker}. While these methods are simple and training-free, they do not fundamentally enhance the model's intrinsic capabilities.

Meanwhile, some pioneer preference optimization methods 
like Direct Preference Optimization (DPO) \cite{rafailov2024direct}
have been introduced, which encourage the model to learn from the comparisons between positive and negative samples, alleviating hallucinations \cite{zhao2023beyond,pi2025strengthening}.
However, most current methods cannot deliver consistent improvements across all types of MLLM hallucination tasks (e.g., VQA and captioning tasks, as shown in our experiments of Table \ref{tab:main}). 
Additionally, it appears that the model's improvement on specific tasks is closely related to the format of the training data.
For instance, the DPO data of SeVa \cite{zhu2024self} primarily consists of VQA, which explains its strong performance on VQA-related hallucination evaluation. However, its results on captioning tasks are relatively unsatisfactory.
Moreover, these methods do not explicitly consider diverse sources of M-hallu.
Hence, we argue that if we focus on mitigating multimodal hallucinations, we should be able to address diverse types of hallucination causes and tasks, and design hallucination-targeted preference pairs for DPO accordingly. Our goal is to comprehensively alleviate all multimodal hallucination problems, including both discriminative tasks (e.g., VQA) and generative tasks (e.g., image captioning).

Different from the hallucinations in LLMs, M-hallu primarily arises from the following three aspects:
(1) \textbf{\emph{Insufficient visual capability}}: This occurs when the MLLM's visual encoder lacks the necessary strength, being distracted by relatively unimportant visual information, leading to hallucinations;
(2) \textbf{\emph{Incapable long-context generation}}: We observe that hallucinations become more pronounced as the generated content grows longer, similar to long-range forgetting, which needs to be addressed in practical applications;
(3) \textbf{\emph{Multimodal conflicts}}: Multimodal conflicts frequently arise in real-world scenarios due to the inevitable noises in texts and images. MLLMs are more prone to hallucinations with conflicting information existing between text and image \cite{liu2024phd}.

To address the aforementioned challenges, we propose \textbf{Hallucination-targeted Direct Preference Optimization (\MODEL)} to mitigate M-hallu.
Our approach constructs hallucination-targeted preference pairs, specifically designed to address various forms and causes of hallucinations.
Specifically, we design three types of DPO data reflecting the corresponding hallucination causes as follows: 
(1) For \emph{insufficient visual capability}, during the model's autoregressive decoding, we preserve only some visual tokens with the lowest attention scores to produce targeted negative responses that reflect incorrect visual information distraction, urging MLLMs to pay attention to more effective visual information.
(2) For \emph{incapable long context generation},
we specifically select positive examples from high-quality long-form captions, while
creating negative examples where the latter part of the response deviates from the image content, simulating long-form hallucinations.
(3) For \emph{multimodal conflicts}, 
we add conflicting information with images into prompts to generate negative examples. We provide positive and negative pairs with questions featuring conflicting prefixes to train the model to correctly respond to the question even containing conflicting information.

We conduct extensive experiments to evaluate our approach across various types of M-hallu tasks. The results demonstrate that our HDPO framework achieves the overall best performance in effectively mitigating MLLM hallucinations on various tasks. Our contributions are summarized as follows:
\begin{itemize}
\item We analyze three key causes behind MLLM hallucinations from visual capability, long-context generation, and multimodal conflicts aspects, offering valuable insights to guide future advancements.
\item Based on these analyses, we propose a novel HDPO, aiming to jointly address all types of M-hallu tasks. To the best of our knowledge, we are the first to adopt hallucination-targeted DPO from diverse aspects with our novel DPO data construction strategies.
\item Through extensive experiments on different datasets, HDPO demonstrates consistent improvements in all types of M-hallu tasks.
\end{itemize}

\section{Related Work}
\subsection{Mitigating Hallucinations in Multimodal Large Language Models}

Recently, lots of works have explored various approaches to mitigating M-hallu \cite{liu2023mitigating, chen2024halc, lee-etal-2024-volcano, yu2024rlhf, wu-etal-2024-logical}, showing promising results and offering valuable insights.\\
\noindent\textbf{Training Free Methods.} 
VCD \cite{leng2024mitigating} mitigates M-hallu by reducing model's knowledge bias through contrastive decoding, which is effective but requires model to inference twice for each token prediction, resulting in higher memory consumption and increased overhead in real-world applications.
OPERA \cite{huang2024opera} tackles hallucinations by identifying some common patterns of decoding attention scores when model hallucinates and applying special decoding strategies,
while these measures increase inference load and slow down processing speed.
WoodPecker \cite{yin2023woodpecker} utilizes external feedback to reduce hallucinations, but its reliance on external tools adds complexity without enhancing the model’s intrinsic capabilities.
Some works further focus on the internal state of MLLMs to discover their abnormalities \cite{zhang2024pip}.
To seek improvement in the intrinsic capabilities of models, we turn to training paradigms as our approach.

\noindent\textbf{DPO methods for improving MLLM.}
HA-DPO \cite{zhao2023beyond} views hallucinations as models' preferences.
By leveraging ChatGPT \cite{achiam2023gpt} alongside ground truth annotations from existing image datasets, it generates positive examples aligned with image content, while the model’s original outputs serve as negative examples for direct preferences optimization.
Although effective, the construction of negative examples is suboptimal, as it may not fully capture the diverse forms of M-hallu.
SeVa \cite{zhu2024self} generates negative examples by introducing noise into images and treats the model's original outputs as positive examples, constructing pairs for DPO.
In addition to adding noise, BPO \cite{pi2025strengthening} injects errors into positive examples via the LLM backbone of MLLMs to construct negative examples.
However, our experiments indicate that while these methods demonstrate strong capabilities, their performance in hallucination-related evaluations is not particularly impressive.
Nonetheless, these works demonstrate the superiority of DPO in enhancing models' capabilities.
Inspired by these findings, we aim to develop methods to further mitigate M-hallu from its diverse forms with hallucination-targeted direct preference optimization.

\noindent\textbf{\MODEL differs from existing methods.}
Unlike other existing preference optimization approaches, we primarily focus on hallucination-targeted preference optimization.
We analyze and address hallucinations in MLLMs from diverse causes and forms.
During the preference optimization process, the model learns to distinguish between positive and negative examples.
HA-DPO enables the model to be aware of hallucinated content in its original outputs, though its effectiveness is limited to capturing the diverse range of hallucinations as the data is insufficient. %
In contrast, other works use general preference data, which improves overall model capability but shows inconsistency across different hallucination benchmarks.
Therefore, we aim to enhance the effectiveness of DPO by constructing examples that reflect a wider range of hallucination forms and characteristics, allowing the model to align better to make less hallucination.

\subsection{Causes of Hallucinations in Multimodal Large Language Models}
There are substantial works exploring M-hallu, offering insightful perspectives.
VCD suggests that language prior within MLLM is a key factor in inducing hallucinations.
The Less is More \cite{yue-etal-2024-less} highlights that hallucinations are more prevalent in longer texts.
In contrast, Eyes Wide Shut \cite{tong2024eyes} identifies limitations in the current CLIP-based visual encoders used in MLLMs, showing that they fail to capture fine-grained details.
Furthermore, SID \cite{huo2024self} points out that tokens with lower weights in the early layers can trigger subsequent hallucinations.
Meanwhile, PhD \cite{liu2024phd} demonstrates that M-hallu stems from conflicts between multimodal information, and counterintuitive images particularly prone to causing hallucinations.
Collectively, these studies provide valuable insights into understanding and addressing M-hallu.

\section{Method}
In this section, we provide a brief preliminaries of MLLM and DPO, followed by a detailed explanation of our proposed method for constructing three types of hallucination-targeted preference data.
\begin{figure*}[h]
    \centering
    \includegraphics[width=\linewidth]{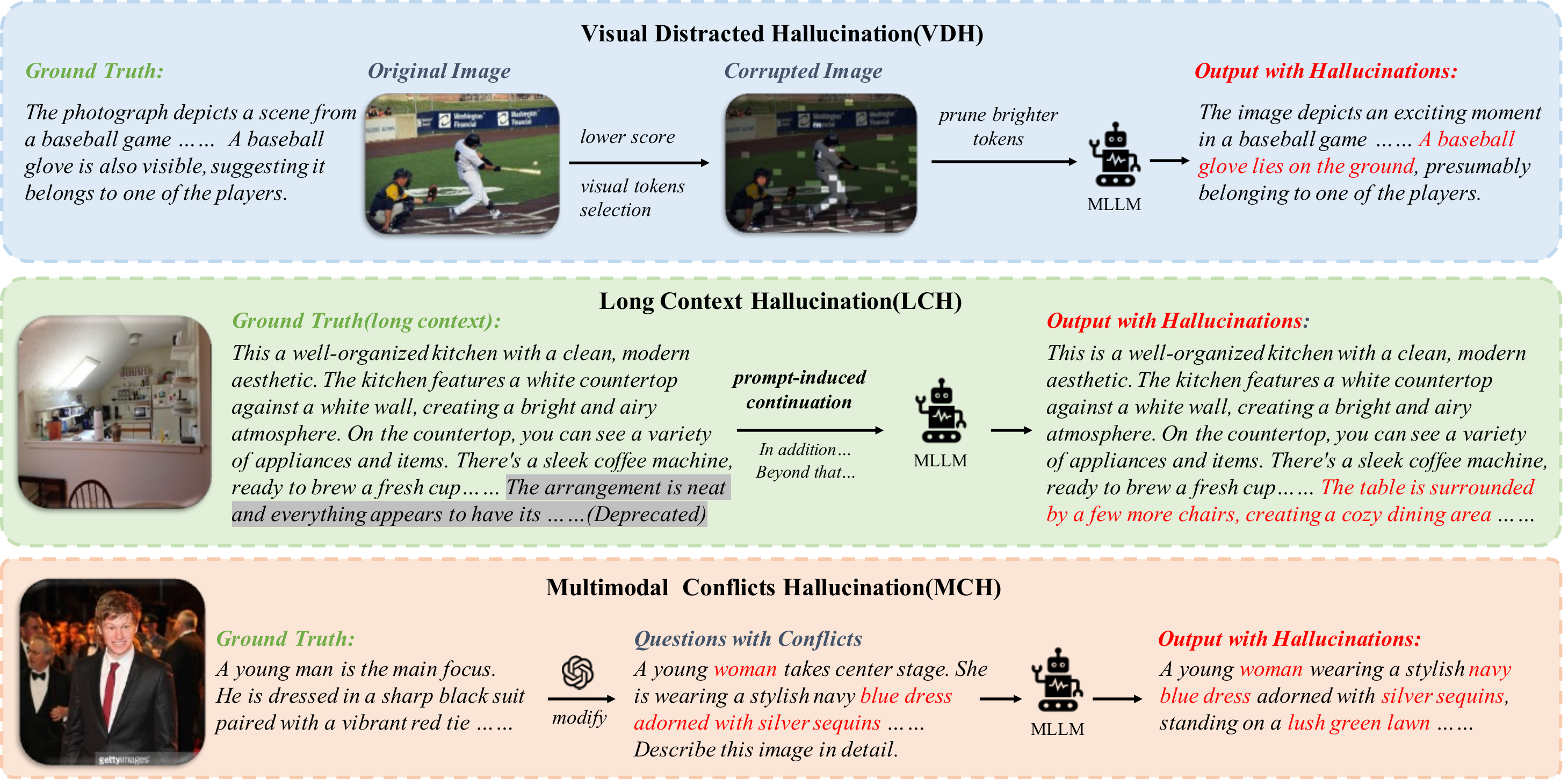}
    \vspace{-1.5em}
    \caption{Overview of our three kinds of Hallucinated-targeted Preference data. Better view on the digital screen.}
    \vspace{-1.0em}
    \label{fig:overview}
\end{figure*}
\subsection{Preliminaries}

\textbf{Multimodal Large Language Models.} 
MLLMs utilize LLMs to predict the probability distribution of the next token for each textual input.
Given a prompt $x$ that includes both an image and a text query, MLLMs generate a corresponding text response $y$.
By incorporating visual information, MLLMs enhance the capabilities of LLMs, enabling multimodal understanding.

\noindent\textbf{Direct Preference Optimization.}
To better align LLMs with human preferences, preference optimization methods have been developed.
Among these, Reinforcement Learning from Human Feedback (RLHF) is a widely recognized method,
though it involves training a reward model, which can be quite challenging.
In contrast, Direct Preference Optimization (DPO) \cite{rafailov2024direct} utilizes preferences data directly, without the need for a reward model.
This makes DPO the approach we employ.
Given a pre-processed preference dataset $D$ containing $x$, $y_c$, and $y_r$, where $x$ represents the input prompt, $y_c$ is the preferred response, and $y_r$ is the rejected response, DPO optimizes the language model through the following loss function:
\begin{equation*}
\resizebox{\linewidth}{!}{$
    \mathcal{L}_{\text{d}} = -\mathbb{E}_{\mathcal{D}}\left[\log \sigma\left(\beta \log\frac{\pi_{\theta}(y_c|x)}{\pi_{\text{ref}}(y_c|x)}- \beta\log\frac{\pi_{\theta}(y_r|x)}{\pi_{\text{ref}}(y_r|x)}\right)\right]
    \label{eq:dpo}
$}
\end{equation*}
where $\pi_{\text{ref}}(y|x)$ denotes the reference policy, i.e., the language model after supervised fine-tuning, with $\theta$ as the trainable parameter.
In our approach, we apply DPO for preference optimization to adjust the MLLM's parameters.

\subsection{Analysis for Construction}

The primary aim of our approach is to tackle the hallucination problem in MLLMs by constructing hallucination-targeted preference pairs, rather than relying on general preference data.
Without loss of generality, we adopt a generalized data format: image-descriptive text data,
which we believe more effectively captures various forms of hallucination.

For DPO in MLLMs, we require a preference dataset $D$, denoted as $(I, x, y_c, y_r)$, where $I$ is the image, $x$ is the question, $y_c$ is the preferred (positive) response, and $y_r$ is the rejected (negative) response.
Currently, there are already many high-quality positive examples available, such as the refined positive examples in HA-DPO for the VG dataset, which leverage ChatGPT to enhance image annotations, and a vast number of positive examples labeled by GPT-4V in ShareGPT4V \cite{chen2023sharegpt4v}. 
These high-quality datasets have a strong alignment with the image content, making them suitable for use as positive examples in DPO.
Therefore, our focus going forward is on how to construct more valuable and informative negative examples, particularly those that target hallucination, which will help the model learn preferences and reduce hallucination occurrences.

To this end, we develop three types of pairwise samples specifically targeting hallucination issues: Visual Distracted Hallucination (\textbf{\A}), Long Context Hallucination (\textbf{\B}) and Multimodal Conflict Hallucination (\textbf{\C}).
An overview of each data type is provided in \cref{fig:overview}, and further details are outlined in the sections below.

\subsection{Sample Construction of Visual Distracted Hallucination}

Inspired by SID, we induce vision-and-text association hallucinations by leveraging vision tokens with low attention scores in the self-attention module.
Formally, for the transformer block in the auto-regressive decoder, text instructions, vision inputs, and generated tokens are concatenated and projected into three vectors: Q, K and V.
The self-attention computes the relevance of each element to the others as follows to get the attention matrix:
\begin{equation}
\begin{split}
\mathbf{A}=\mathrm{softmax}(\frac{\mathbf{Q}\cdot \mathbf{K}^{T}}{\sqrt{d}} + M)\nonumber
\label{eq4}
\end{split}
\end{equation}
where $d$ represents the dimension of \textbf{Q}, \textbf{K}, \textbf{V}, $M$ represents the casual mask. $\mathbf{A} \in R^{(b, h, n, n)}$, where $b$, $h$, and $n$ denote batch size, number of key-value heads, and total token number, respectively.
We denote the $\mathbf{A}_{i}$ as the attention matrix after Layer $i$ of MLLMs.
Then we calculate vision token importance scores ($\mathrm{Score}_{i}(v)$) based on $\mathbf{A}_{i}$:
\begin{equation}
\mathrm{Score}_{i}(v)  = \frac{1}{h}\sum_{j=1}^h\mathbf{A}_{i}^{(\cdot,j,\cdot,\cdot)}[-1]\nonumber
\label{eq5}
\end{equation}
During the model's auto-regressive decoding process, we retain the $K$ vision tokens with the lowest importance scores, and the resulting decoded response serves as negative samples.
By removing the most important visual token from the model in this way, amplifying the influence of relatively irrelevant visual tokens, thus constructing visual information distracted hallucinations
as negative samples, uring MLLMs to pay attention to more important visual information.

\subsection{Sample Construction of Long Context Hallucination}
\begin{figure}[h]
    \centering
    \includegraphics[width=\linewidth]{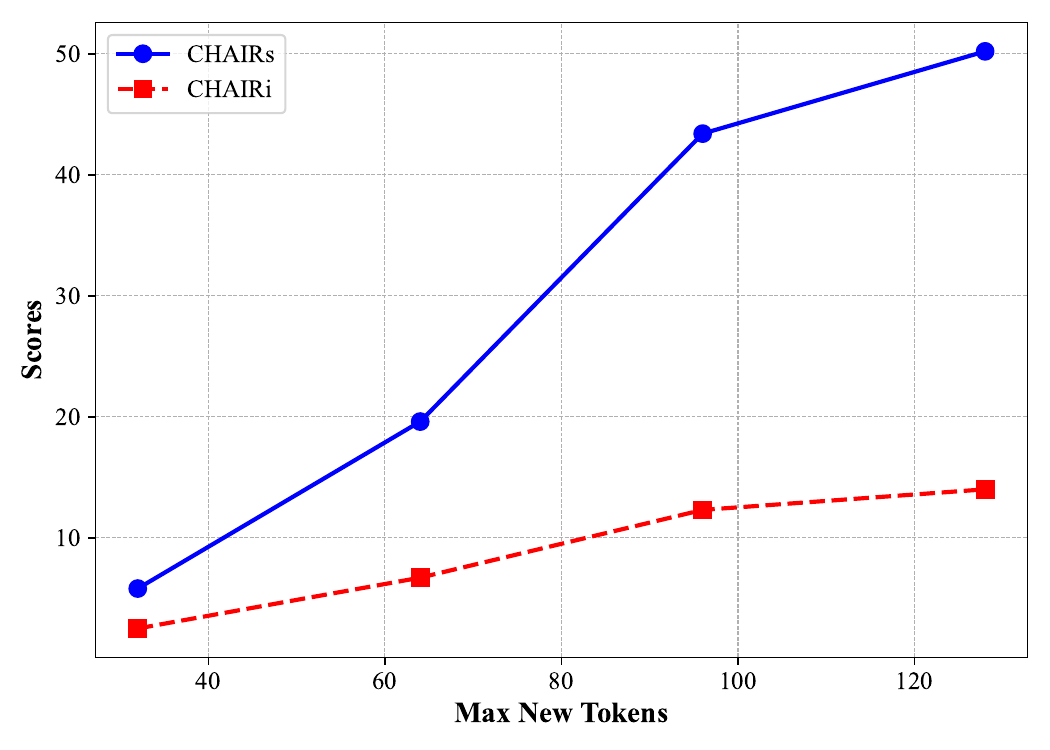} %
    \vspace{-1.5em} %
    \caption{CHAIR scores under different max new tokens}
    \vspace{-1.0em} %
    \label{fig:cs_ci}
\end{figure}

As previously mentioned, the occurrence of hallucinations tends to increase as models generate longer responses. 
To illustrate this more clearly, we present CHAIR scores by varying the '\textit{max new tokens}' parameter. 
As shown in \cref{fig:cs_ci}, the CHAIR score of LLaVA-v1.5-7B exhibits a clear positive correlation with the 'max new tokens', indicating that more hallucinations are produced as the generated content increases.
This issue has also been highlighted in recent studies \cite{yue-etal-2024-less}.

This phenomenon is both logical and explainable.
As the model generates longer texts, the proportion of text tokens gradually increases while the proportion of image tokens decreases.
This shift causes the model to increasingly neglect visual tokens, resulting in descriptions that appear reasonable but fail to accurately align with the visual content.
Our aim is to construct preference data that guides the model to better align its generated content with the visual input and the given question, even when generating long responses.

Given our focus on relatively long-form content, the responses need to be sufficiently lengthy (high-quality long responses).
For negative examples, we truncate the last two sentences from a positive example and use the preceding portion as a prefix.
The MLLM then continues generating text from this prefix, which compels the model to produce common errors associated with extended text generation.
This process is repeated by concatenating the newly generated content to the prefix for three iterations in a loop.

\textbf{Hint Phrase} Simply providing the prefix and instructing the model to continue often results in unexpected behavior, as the model tends to conclude the response quickly, generating low-information descriptions.
To address this issue, we append a 'hint phrase' to the prefix, guiding the model toward producing more informative and detailed responses.
Besides that, we also modify the system prompt. Details can be seen in \cref{long-details}.
It helps produce responses prone to more likely errors when generating long texts. 
By creating positive and negative pairs in this manner, we aim to use DPO to teach the model how to minimize hallucinations in long-form responses and improve alignment.

\subsection{Sample Construction of Multimodal Conflicts Hallucination}
One of the more challenging yet often overlooked scenarios in mainstream evaluation tasks involves conflicts between modalities.
In such cases, models may naturally favor textual content due to their autoregressive generating manner and the larger proportion of the language model component, leading to incorrect outputs.
To address this, we construct positive and negative pairs with conflicting prefixes and apply DPO to optimize the model.

\begin{figure}[h]
    \centering
    \includegraphics[width=\linewidth]{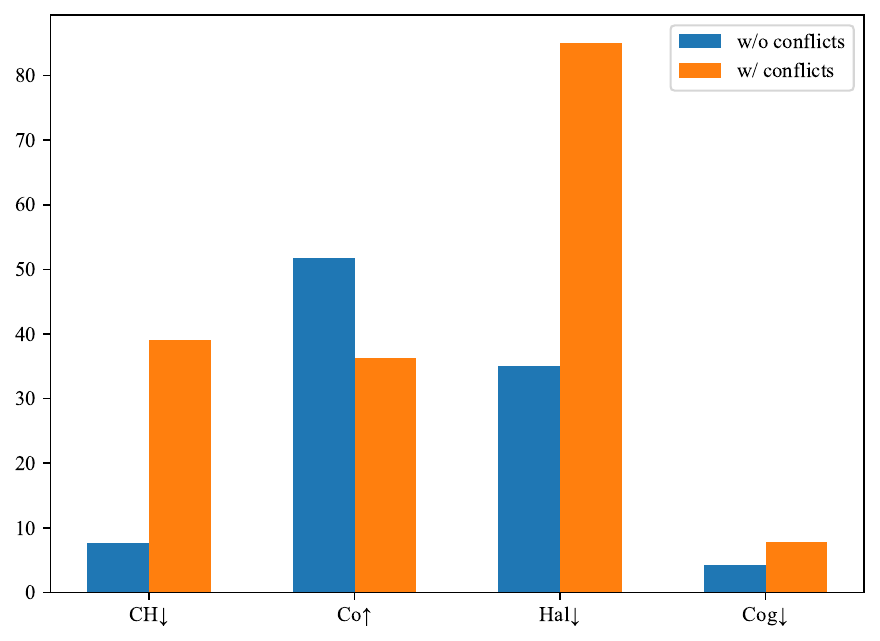} %
    \vspace{-2.0em} %
    \caption{Performance of LLaVA-v1.5-7B w/ and w/o conflicts on AMBER, details in \cref{cf-lab}.}
    \vspace{-1.5em} %
    \label{fig:cf_dec}
\end{figure}

Specifically, we utilize GPT-4o-mini to rewrite details of the positive examples through prompting, generating information conflicting with the image contents.
These conflicting informations are then placed at the beginning of normal questions, 
prompting the model to produce incorrect responses.
As shown in \cref{fig:cf_dec}, the model is indeed prone to being hallucinated by the conflicting prefixes.
We take the model's incorrect outputs as negative examples.
Further details on the prompts can be found in \cref{fig:modifyp}. 
Unlike previous types of data, the questions for training of \C contain conflicting prefixes, as we aim for the model to generate correct responses in the query even when presented with conflicting information.
\subsection{Implement details}
For \B, which requires longer responses, we sampled 6k examples with over 300 tokens from ShareGPT4V. For \C, we randomly sampled 6k examples from ShareGPT4V. For \A, we obtain 6k examples from ShareGPT4V and $~$4k examples from VG with positive examples from HA-DPO to enhance data diversity; the preserved K is 500, with other settings aligned with SID (e.g., i = 2). Details of data can be found in \cref{data-details}.

\begin{table*}[t]
    \centering
    \small
    \setlength{\tabcolsep}{3pt} %
    \renewcommand{\arraystretch}{1.2} %
    \begin{tabular}{lrrrrrrrrr}
        \toprule
        & \multicolumn{1}{c}{POPE} & \multicolumn{2}{c}{CHAIR} & \multicolumn{6}{c}{AMBER} \\
        \cmidrule(lr){2-2} \cmidrule(lr){3-4} \cmidrule(lr){5-10}
        & F1 Score $\uparrow$ & CHAIR$_{s}$ $\downarrow$ & CHAIR$_{i}$ $\downarrow$ & CHAIR $\downarrow$ & Cover. $\uparrow$ & HalRate $\downarrow$ & Cog. $\downarrow$ & F1 Score $\uparrow$ & AMBER-S $\uparrow$\\
        \midrule
        LLaVA-v1.5-7B & 86.1 & 51.2 & 14.2 & 7.6 & 51.8 & 35.1 & 4.3 & 74.5 & 83.5 \\
        Vlfeedback $^\dagger$ & 83.7 & 40.3 & 13.2 & --  & --   & --   & --  & --   & --   \\
        POVID $^\dagger$    & 86.9 & 35.2 & 8.3  & --  & --   & --   & --  & --   & --   \\
        CLIP-DPO $^\sharp$    & 85.8 & --   & --   & \underline{3.7} & 47.8 & \underline{16.6} & \underline{1.3} & 82.9 & 89.6 \\
        HA-DPO      & \underline{86.9} & 37.2 & 10.0 & 6.4 & 49.7 & 29.9 & 3.2 & 78.2 & 85.9 \\
        SeVa        & 86.8 & 54.6 & 15.9 & 7.4 & \underline{53.4} & 35.6 & 3.2 & 84.1 & 88.3 \\
        BPO         & 83.1 & 42.2 & 10.1 & 5.0 & \textbf{58.8} & 33.5 & 2.0 & \textbf{84.5} & \underline{89.7} \\
        CSR       & \textbf{87.0} & \underline{19.6} & \underline{5.4}  & 3.8 & 45.0 & 16.9 & 1.4  & 76.0 & 86.1 \\ \midrule
        \MODEL(ours)& 86.8 & \textbf{16.6} & \textbf{5.1} & \textbf{3.3} & 50.2 & \textbf{15.8} & \textbf{0.8} & \underline{84.1} & \textbf{90.4}\\
        \bottomrule
    \end{tabular}
    \caption{Experimental results of LLaVA-v1.5-7B trained with our hallucination-targeted direct preference optimization compared with baselines applied on LLaVA-v1.5-7B.
    The best result for each metric is in bold, and the second-best result is underlined. Some results are referenced from: $^\dagger$\citet{zhou2024calibrated}, and $^\sharp$\citet{ouali2024clip}.}
    \label{tab:main}
    \vspace{-1.0em} %
\end{table*}

\section{Experiment}
In this section, we empirically investigate the evaluation of \MODEL.
We begin by describing the experimental settings, including the evaluation datasets used in our study and training details.
Next, we present the results on various hallucination evaluation datasets, demonstrating the promising performance of \MODEL.
Additionally, we validate the expected functions of \B and \C.
Finally, we provide ablation studies and conduct in-depth analyses of the model in more detail.

\subsection{Experimental Settings}
\subsubsection{Datasets and Evaluation Metrics}
We evaluate the effectiveness of \MODEL in mitigating hallucinations across both captioning tasks and simplified visual question answering (VQA) tasks using three evaluation datasets as follows:

\noindent\textbf{CHAIR} \cite{rohrbach-etal-2018-object}
is an evaluation method used in image captioning tasks to assess object hallucinations in model responses.
There are two metrics: CHAIR$_{s}$ and CHAIR$_{i}$.
CHAIR$_{s}$ measures hallucinations at the sentence level, while CHAIR$_{i}$ measures them at the image level respectively.
We conduct the CHAIR evaluation on the MSCOCO dataset following the setting in OPERA with 500 random images.
For each image, the model is prompted with: "Please describe this image in detail." to obtain their descriptions.
By default, we set the max new token to 512.
More specifically, the calculation for the CHAIR$_{s}$  and CHAIR$_{i}$ metrics are as follows:
\begin{equation}
\text{CHAIR$_{s}$} = \frac{|\{\text{hallucinated objects}\}|}{|\{\text{all mentioned objects}\}|}
\nonumber
\end{equation}
\begin{equation}
\text{CHAIR$_{i}$} = \frac{|\{\text{captions w/ hallucinated objects}\}|}{|\{\text{all captions}\}|}
\nonumber
\end{equation} 

\noindent\textbf{POPE} \cite{li-etal-2023-evaluating}
is a popular dataset for evaluating object hallucinations in MLLMs.
The evaluation is asking the model questions in the format: "Is there a <object> in the image?".
It can be divided into three splits: popular, adversarial, and random.
We evaluate the metrics for all splits, calculate and report the average F1 score.

\noindent\textbf{AMBER} \cite{wang2023llm}
is an LLM-free multi-dimensional benchmark, offering a cost-effective and efficient evaluation pipeline.
It supports the evaluation of both generative and discriminative tasks including hallucinations related to existence, attributes, and relations.
For its generative task, four metrics are used: CHAIR, Cover, Hal and Cog.
For its discriminative task, we calculate and report the average F1 score.
We also calculate AMBER Score denoted as AMBER-S, which reflects overall performance.
Details of metrics can be found in \cref{ab}.

\subsubsection{Training details}
Following previous works \cite{chen2023sharegpt4v,zhu2024self,pi2025strengthening}, we select the LLaVA-v1.5-7B \cite{NEURIPS2023_6dcf277e} and LLaVA-v1.5-13B as base models for experiments, which allows for easy comparison with other existing works.
The LLaVA's weights are pretrained and further fine-tuned using supervised fine-tuning (SFT) before applying our hallucination-targeted direct preference optimization.
During the training phase, we employ Zero stage-3 optimization and use Vicuna-7B/13B and CLIP-VIT-L-336px as our LLM and vision encoder, respectively.
The training is conducted with 2 epochs with a batch size of 64, a learning rate of 2e-6, weight decay as 0, LoRA rank as 64, and a beta value of 0.1.
All experiments are run on one single machine with 8 A800 GPUs.
The total training time is 3 hours for LLaVA-v1.5-7B and 4 hours for LLaVA-v1.5-13B.
\subsection{Results on Hallucination Evaluation}
We compare the performance of \MODEL with several baselines and the experimental results are shown in \cref{tab:main}.
We selected Vlfeedback\cite{li-etal-2024-vlfeedback}, POVID\cite{zhou2024aligning}, CLIP-DPO\cite{ouali2024clip}, HA-DPO, SeVa, BPO, and CSR\cite{zhou2024calibrated} as our baselines.
We can observe and analyze as follows:

\noindent\textbf{\MODEL achieves SOTA level on hallucination.}
The results indicate that \MODEL performs well in mitigating hallucinations, achieving almost SOTA level, especially on generative tasks.
This outcome is natural, as our data contains only descriptive content, leading to relatively strong performance on generative tasks.
However, despite not using VQA-type data, \MODEL still performs well on VQA tasks.
This demonstrates that our approach addresses the fundamental issue of hallucinations in the model, rather than merely excelling in specific task types.
While there is a slight decrease on AMBER Cover. metric, we believe this is due to an inherent trade-off between achieving more complete object coverage and reducing hallucination rates.
Notably, \MODEL is able to reduce the HalRate from 35.1\% to 15.8\%, representing an improvement of nearly 55\%, with only a minimal drop in Cover. compared to the original model, which is delighted to accept.
In fact, striking the optimal balance between these metrics remains an open question and lies outside the scope of this paper.

We also evaluate \MODEL on a comprehensive benchmark, MM-Vet \cite{yu2024mm}, where we observe a slight improvement.
This aligns with our expectations, as the model is not fine-tuned on a wide range of tasks and data types, but instead focused specifically on reducing hallucinations.

\noindent\textbf{Brief analysis on other baselines.}
Some baselines lack comprehensive performance on hallucination evaluation.
SeVa, though effective on AMBER's discriminative tasks, shows no improvement on generative tasks, likely due to its reliance on VQA-type data.
Similarly, BPO underperforms on CHAIR.
In contrast, CSR excels in generative tasks but struggles with AMBER’s discriminative tasks.
This indicates that while these methods enhance model performance, they do not fully optimize for hallucination, and their ability to mitigate hallucinations remains inconsistent and incomplete, while \MODEL demonstrates strong performance in hallucination evaluation, as evidence of its 'hallucination-targeted' design.

\noindent\textbf{Advantage on DPO Data Efficiency.}
The size of our dataset also provides a relative advantage.
For instance, BPO fine-tunes the model using nearly 180,000 examples.
In contrast, with only around 22,000 examples, \MODEL significantly improves model’s performance on hallucination. Detailed experiments are given in Section \ref{sec:scaling}.

\begin{table*}[t]
    \centering
    \small
    \setlength{\tabcolsep}{3pt} %
    \renewcommand{\arraystretch}{1.2} %
    \begin{tabular}{lrrrrrrrrr}
        \toprule
        & \multicolumn{1}{c}{POPE} & \multicolumn{2}{c}{CHAIR} & \multicolumn{6}{c}{AMBER} \\
        \cmidrule(lr){2-2} \cmidrule(lr){3-4} \cmidrule(lr){5-10}
        & F1 Score $\uparrow$ & CHAIR$_{s}$ $\downarrow$ & CHAIR$_{i}$ $\downarrow$ & CHAIR $\downarrow$ & Cover. $\uparrow$ & HalRate $\downarrow$ & Cog. $\downarrow$ & F1 Score $\uparrow$ & AMBER-S $\uparrow$\\
        \midrule
        LLaVA-v1.5-13B & 85.8  & 48.0 & 13.6 & 6.6 & 52.0 & 31.0 & 3.3 & 73.0 & 83.2\\
        HA-DPO         & \underline{87.3} & 46.0 & 12.1 & 6.0 & \underline{52.3} & 30.7 & 3.0 & 79.1 & 86.6 \\
        SeVa           & 86.9 & 59.8 & 17.4 & 9.0 & \textbf{54.5} & 43.3 & 3.7 & \textbf{84.8} & \underline{87.9} \\
        CSR            & \underline{87.3} & \underline{24.0} & \underline{5.6}  & \textbf{3.6} & 48.1 & \underline{19.0} & \underline{1.8} & 73.1 & 84.8 \\ \midrule
        \MODEL(ours) & \textbf{87.6} & \textbf{15.4} & \textbf{5.3} & \underline{3.8} & 51.0 & \textbf{16.5} & \textbf{0.8} & \underline{81.2} & \textbf{88.7}\\
        \bottomrule
    \end{tabular}
    \caption{Experimental results of LLaVA-v1.5-13B trained with hallucination-targeted direct preference optimization compared with baselines applied on LLaVA-v1.5-13B. 
    The best result for each metric is in bold, and the second-best result is underlined.}
    \label{tab:13b}
    \vspace{-1.0em} %
\end{table*}

\subsection{Results on Different Base LLMs}

We also conduct experiments across base LLMs of different sizes to verify our HDPO's universality.
Specifically, we apply \MODEL to the widely-used LLaVA-v1.5-13B for MLLM hallucination evaluation.
The results are shown in \cref{tab:13b}, demonstrating that the model's performance remains consistent with expectations, with improvements in hallucination mitigation.
It also implies that our generated hallucination-targeted DPO data is effective for different LLM sizes.

\subsection{In-Depth Model Analyses}
We propose our method for generating different types of hallucination-based preference pairs.
The results from our main experiment demonstrate our method's superior performance in mitigating hallucinations.
However, do they truly work effectively in the scenarios we claim?
Below, we briefly design two more challenging sub-tasks of hallucination that align with our claims, aiming to further showcase the effectiveness of our data construction of \B and \C.
\subsubsection{Longer Description Experiments}
To evaluate the effectiveness of \B on longer responses, we have conducted an extended experiment on the AMBER generative task.
Specifically, when the model is asked the question "Describe this image in detail", we append the instruction "answer in 800 words" to encourage longer responses.
As indicated in \cref{tab:long}, \MODEL shows good and stable performance in handling longer responses, with the lowest HalRate, CHAIR$_s$, and Cog.
Although BPO has the highest Cover, its HalRate is quite poor.
Our \MODEL, compared to the original model, maintains a similar level of Cover while significantly reducing the HalRate.
It demonstrates that our construction for \B works as expected in mitigating longer responses.
\begin{table}[t]
    \centering
    \small
    \setlength{\tabcolsep}{4pt} %
    \renewcommand{\arraystretch}{1.2} %
    \begin{tabular}{lrrrr}
        \toprule
         & CHAIR$_{s}$ $\downarrow$ & Cover. $\uparrow$ & HalRate $\downarrow$ & Cog. $\downarrow$ \\
        \midrule
        LLaVA-1.5-7B  & 9.0 & 54.4 & 45.1 & 5.7 \\
        HA-DPO      & 7.5 & 52.9 & \underline{37.6} & 4.4 \\ 
        SeVa        & 7.5 & \underline{57.8} & 43.4 & \underline{4.3} \\
        BPO         & \underline{6.4} & \textbf{65.1} & 55.3 &	4.8 \\ 
        \MODEL      & \textbf{3.4} & 54.7 & \textbf{21.4} & \textbf{1.3} \\
        \bottomrule
    \end{tabular}
    \caption{Results of longer description experiments.}
    \vspace{-1.0 em}
    \label{tab:long}
\end{table}

\subsubsection{Multimodal Conflicts Experiments}
\label{cf-lab}
In real-world scenarios, multimodal conflicts are common when using MLLMs.
To better evaluate the model's performance under such conditions, we design a more challenging task.
Specifically, we randomly select 200 questions from the generative task in the AMBER dataset.
First, LLaVA-1.5-7B is used to generate answers for these questions to get coarse-grained image descriptions.
Next, GPT-4o-mini rewrites the details in the descriptions, following the construction method of \C.
We then introduce the incorrect information as a prefix to the question and ask the model to describe the image while influenced by the conflicting context.

The experimental results are shown in \cref{tab:conf}, demonstrating that despite encountering conflicting prefixes, our \MODEL maintains promising performance.
Compared to other baselines, \MODEL achieves the best scores in CHAIR$_s$, HalRate, and Cognition, along with improved coverage over the original model, which aligns with our expectations.
It reveals that our \MODEL shows significant improvement in the model's performance under this more difficult setting, highlighting the effectiveness of \C.
Additionally, we also make comparison between the effects of adding noise and preserved visual tokens with lower scores. Further details can be seen in the \cref{comp}.
\begin{table}[t]
    \centering
    \small
    \setlength{\tabcolsep}{4pt} %
    \renewcommand{\arraystretch}{1.2} %
    \begin{tabular}{lrrrr}
        \toprule
        & CHAIR$_{s}$ $\downarrow$ & Cover. $\uparrow$ & HalRate $\downarrow$ & Cog. $\downarrow$ \\
        \midrule
        LLaVA-1.5-7B & 39.1 & 36.3 & 85.1 & 7.8 \\
        HA-DPO     & 40.3 & 35.7 & 86.1 & 8.1 \\
        SeVa       & 39.1 & 36.5 & 86.1 & 7.8 \\
        BPO        &\underline{22.3} & \textbf{56.0} & \underline{81.2} & \underline{7.7} \\
        \MODEL     & \textbf{14.3} & \underline{49.5} & \textbf{52.0} & \textbf{5.2} \\ 
        \bottomrule
    \end{tabular}
    \caption{Results of multimodal conflict experiments.}
    \vspace{-0.8em} %
    \label{tab:conf}
\end{table}

\subsection{Ablation Study}
To demonstrate the contributions of \A, \B, and \C to overall performance, we progressively remove each component and report the results.
As shown in \cref{tab:ablation}, the model's performance declined as we removed each data type.
The model achieves the best performance when all three data types are included.
These experimental results confirm the individual contributions of each component.

\begin{table}[t]
    \centering
    \small
    \resizebox{1.0\linewidth}{!}{
    \renewcommand{\arraystretch}{1.3} %
    \begin{tabular}{lrrrr}
        \toprule
        & \multicolumn{2}{c}{CHAIR} & \multicolumn{2}{c}{AMBER} \\
        \cmidrule(lr){2-3} \cmidrule(lr){4-5}
        & CHAIR$_s$ $\downarrow$ & CHAIR$_i$$\downarrow$ & CHAIR $\downarrow$& F1 $\uparrow$\\
        \midrule
        LLaVA-1.5-7B       & 51.4 & 14.2 & 7.6 & 74.5 \\
        + \A + \B + \C     & \textbf{16.6} &\textbf{ 5.1} & \textbf{3.3} & \textbf{84.1} \\ 
        + \B + \C          & 28.4 & 7.5  & 4.8 & 78.9 \\ 
        + \C               & 51.2 & 15.1 & 7.6 & 78.1 \\
        \bottomrule
    \end{tabular}
    }
    \caption{Impact of progressively removing each data.}
    \label{tab:ablation}
    \vspace{-2.0 em}
\end{table}

\subsection{Scaling Law of \MODEL}
\label{sec:scaling}

\begin{figure}[h]
    \centering
    \vspace{-0.5em}
    \begin{subfigure}[b]{0.49\linewidth}
        \centering
        \includegraphics[width=\linewidth]{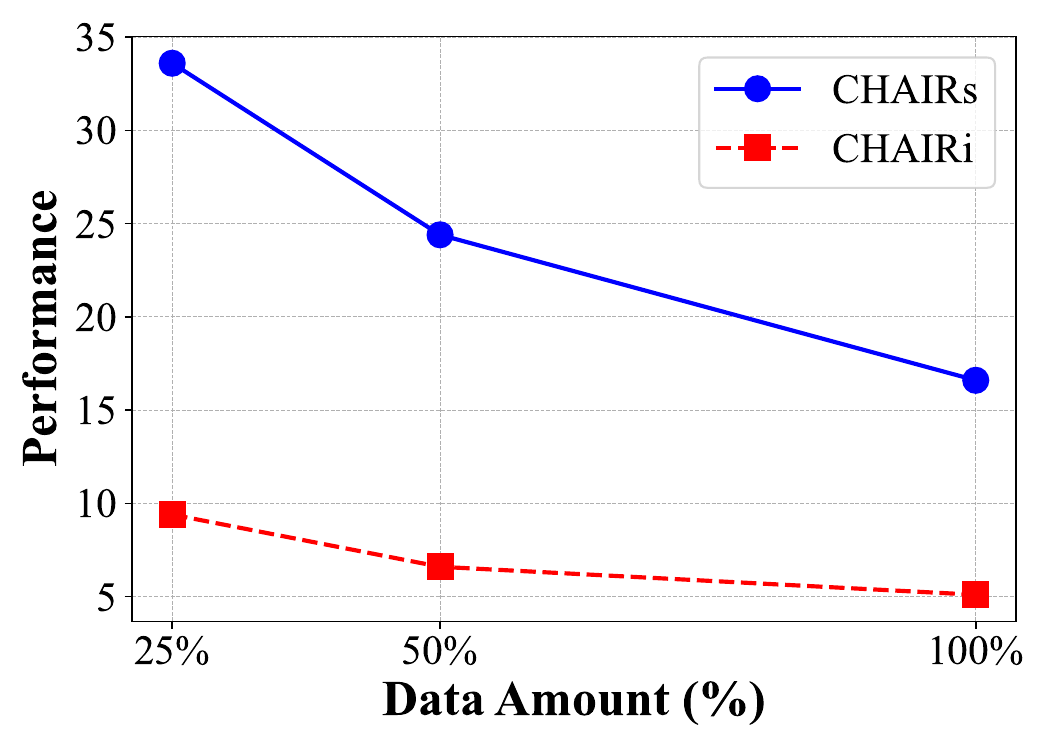} %
        \label{fig:scaling1}
    \end{subfigure}
    \hfill
    \begin{subfigure}[b]{0.49\linewidth}
        \centering
        \includegraphics[width=\linewidth]{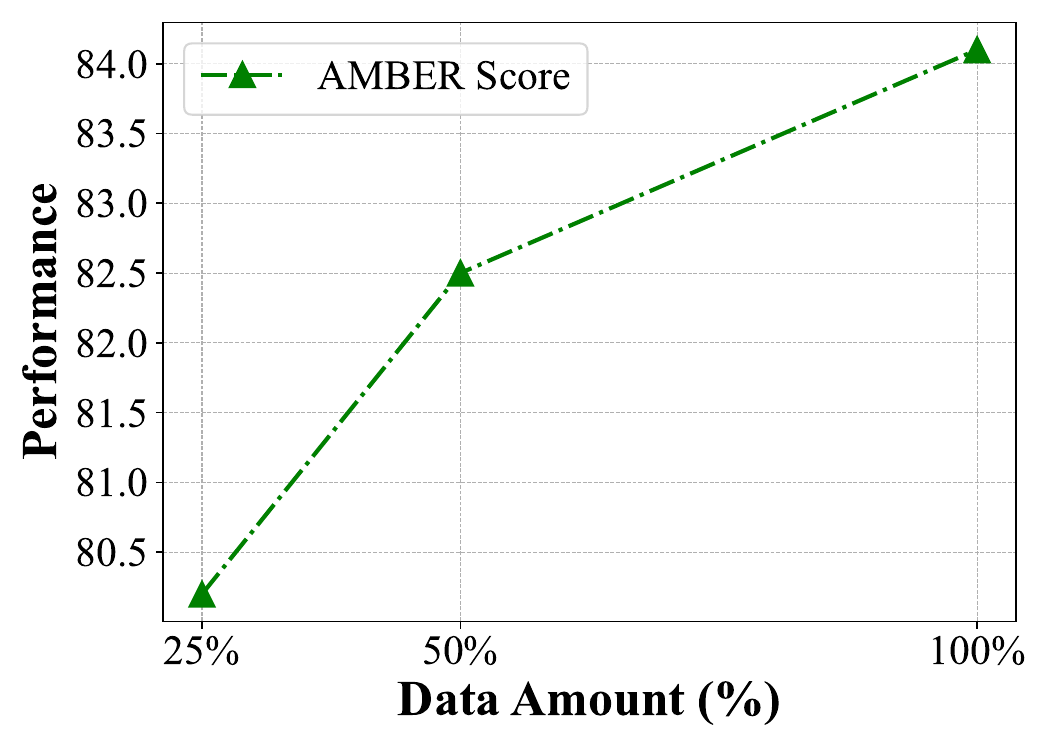} %
        \label{fig:scaling2}
    \end{subfigure}
    \vspace{-2.5em} %
    \caption{Scaling law in \MODEL with different data sizes. Lower CHAIR$_s$ and CHAIR$_i$ are better, while higher AMBER Score is better.}
    \vspace{-1.0em} %
    \label{fig:scaling}
\end{figure}

We analyze the impact of data size on our method.
The performance of LLaVA-v1.5-7B fine-tuned on datasets of varying sizes but the same proportions are shown in \cref{fig:scaling}.
As the data size increases, the effectiveness of our approach also improves, highlighting the potential for scaling up.
This demonstrates the superior performance of \MODEL.

\section{Conclusion}
In this paper, we present \MODEL, a novel approach designed to effectively mitigate hallucinations in MLLMs.
We analyze three types of hallucinations observed in MLLMs and create hallucination preference data based on the identified causes.
Extensive experiments across different benchmarks demonstrate the ability of \MODEL to reduce hallucinations in MLLMs, showing effectiveness.

\section*{Limitations}
In this paper, we introduce \MODEL, which effectively mitigates the hallucination problem in current multimodal large language models. However, several issues remain unresolved.
Specifically, we have not yet developed distinct strategies for controlling data quality, and the generation of automated negative examples lacks methods for further verification and optimization, which could improve the effectiveness of our approach.
Additionally, there may be opportunities to further enhance the quality of positive examples.
Moreover, our construction methods and strategies could potentially be integrated with other techniques for processing more high-quality preference data, which may further improve the model's performance.
Fine-tuning larger models with extensive, integrated datasets may not only enhance overall reasoning capabilities but also increase the model's robustness against hallucinations.
This represents a promising area for further investigation, and we leave these open questions for future research.

\section*{Ethics Statement}
This work mitigates hallucinations of multimodal large language models to enhance their reliability and practicality.
We have carefully considered the ethical implications of our work.
The models and datasets we used are publicly available and commonly used, and our findings may inherit the biases and limitations carried out in these resources.

\section*{Acknowledgements}
We appreciate the valuable discussion with Qidong Huang on conducting experiments, and thanks to him for his helpful suggestions.

\bibliography{custom}
\appendix

\newpage

\label{sec:appendix}

\section{Details of AMBER's Metrics}

\label{ab}
An Automated Multi-dimensional Benchmark for Multi-modal Hallucination Evaluation (AMBER) is an LLM-free multi-dimensional benchmark, offering a cost-effective and efficient evaluation pipeline.
It can be used to evaluate both generative and discriminative tasks including hallucinations related to existence, attributes, and relations.
Its generative evaluation aligns with our desired assessment of long descriptions, while the other dimensions provide insights into the model's performance on relatively simple VQA tasks, thereby reflecting the model's hallucination comprehensively.
For its generative task, four metrics are used: CHAIR, Cover, Hal and Cog.
For its discriminative task, we calculate and report the average F1 score.
Additionally, we also calculate AMBER Score denoted as AMBER-S, which reflects overall performance.

\textbf{CHAIR} measures the frequency of hallucinatory objects in the responses, \textbf{Cover} evaluates the object coverage, \textbf{Hal} represents the proportion of responses containing hallucinations, and \textbf{Cog} assesses whether the hallucinations produced by MLLMs resemble those found in human cognition. \textbf{AMBER Score} is calculated as follows:
\begin{equation}
    \text{\textit{AMBER Score}} = \frac{1}{2}\times(1 - \textit{CHAIR} + \textit{F1})\nonumber
\end{equation}

\begin{table*}[t]
    \centering
    \small
    \setlength{\tabcolsep}{3pt} %
    \renewcommand{\arraystretch}{1.2} %
    \begin{tabular}{lrrrrrrrrr}
        \toprule
        & \multicolumn{1}{c}{POPE} & \multicolumn{2}{c}{CHAIR} & \multicolumn{6}{c}{AMBER} \\
        \cmidrule(lr){2-2} \cmidrule(lr){3-4} \cmidrule(lr){5-10}
        & F1 Score $\uparrow$ & CHAIR$_{s}$ $\downarrow$ & CHAIR$_{i}$ $\downarrow$ & CHAIR $\downarrow$ & Cover. $\uparrow$ & HalRate $\downarrow$ & Cog. $\downarrow$ & F1 Score $\uparrow$ & AMBER-S $\uparrow$\\
        \midrule
        LLaVA-v1.5-7B   & 86.1 & 51.2 & 14.2 & 7.6 & 51.8 & 35.1 & 4.3 & 74.5 & 83.5 \\
        + Diffu$_{6k}$  & 86.2 & 62.8 & 18.4 & 9.2 & \textbf{58.7} & 47.5 & 4.3 & 78.1 & 84.5 \\
        + \A$_{6k}$     & \textbf{87.1} & \textbf{48.2} & \textbf{13.7} & \textbf{6.1} & 57.3 & \textbf{32.0} & \textbf{ 2.7} & \textbf{80.2} & \textbf{87.1} \\
        \bottomrule
    \end{tabular}
    \caption{Experimental results of LLaVA-v1.5-7B trained with two ways to construct preference pairs: adding noise and preserving visual tokens. The diffusion noise step is 800.
    The best result for each metric is in bold.}
    \label{tab:diffu_comp}
    \vspace{-0.5em} %
\end{table*}

\section{Comparison of noise and token preservation}
\label{comp}
We also conduct experiments to compare the impact of adding noise versus preserving visual tokens.
Specifically, we use 6k samples from ShareGPT4V to construct negative samples by introducing diffusion noise and preserving visual tokens, and train the LLaVA-v1.5-7B model by direct preference optimization.
The results of these experiments are presented in \cref{tab:diffu_comp}.
As the experimental results show, using visual token preservation can achieve better performance on hallucination evaluation.

\section{Baseline Selection of 13B}
For the experiments on the 13B model, we select several recent strong baselines, including SeVa and CSR, using their open-sourced checkpoints for evaluation.
Additionally, we reimplement HA-DPO on LLaVA-v1.5-13B, as the original repository does not provide this checkpoint.
We also attempt to reimplement BPO on LLaVA-13B with no available checkpoints, the evaluation results are unexpectedly low, with POPE scores falling below 80.0.
Therefore, these results are not included in the table.
However, the BPO results for the 7B model are obtained using the publicly released checkpoints.
For CLIP-DPO, we are unable to find the official repository, and thus, it is not compared here.

\section{Details about Our data}
\label{data-details}

\subsection{Visual Disctracted Hallucination}
\begin{figure}[h]
    \centering
    \includegraphics[width=\linewidth]{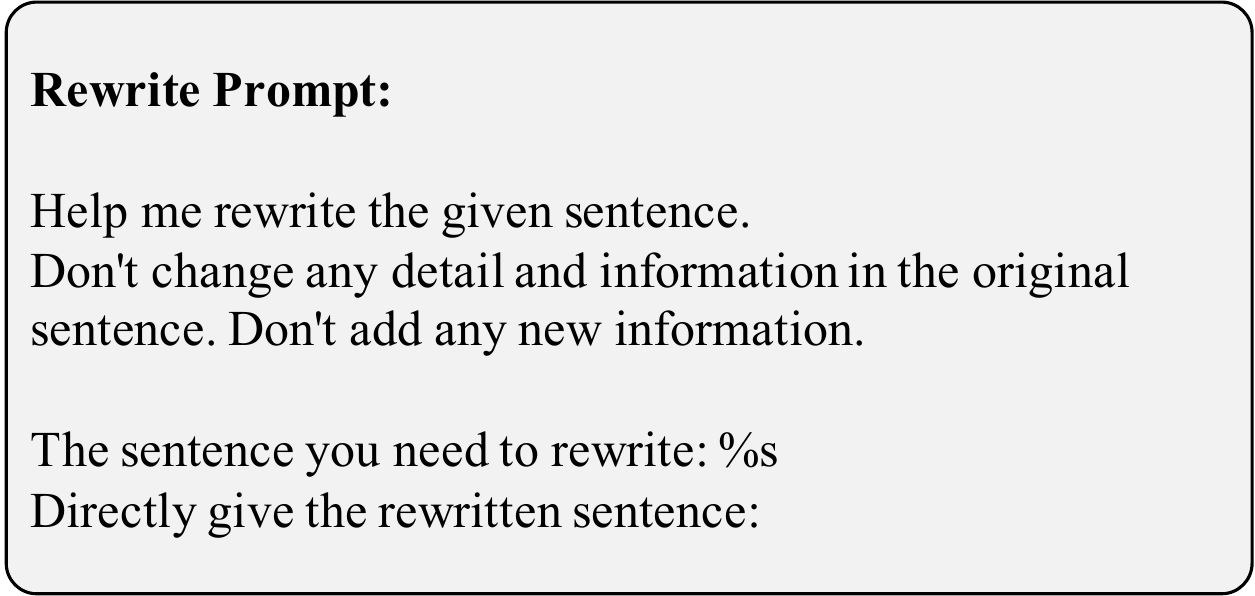}
    \vspace{-1.8em}
    \caption{Rewrite Prompt used in \A}
    \vspace{-0.5em}
    \label{fig:rewritep}
\end{figure}

\begin{figure}[h]
    \centering
    \includegraphics[width=\linewidth]{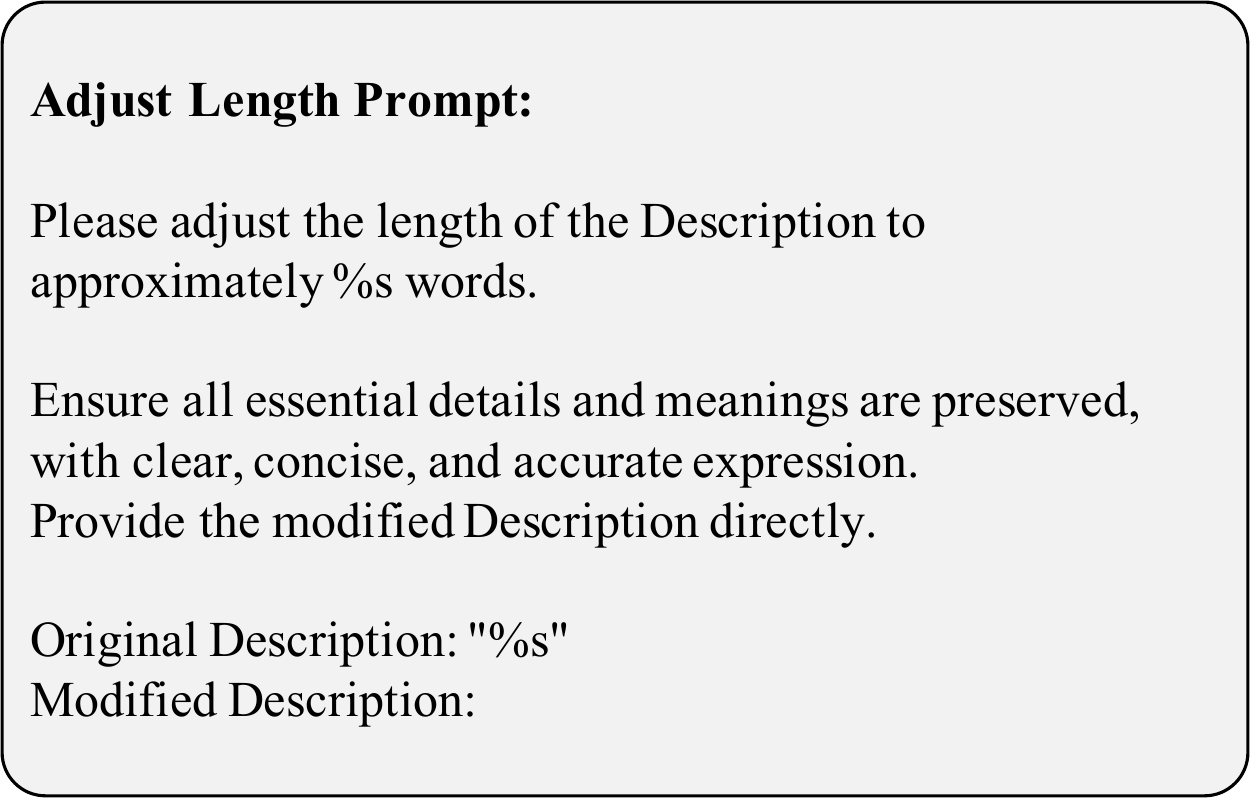}
    \vspace{-1.8em}
    \caption{Adjust Length Prompt used in \A}
    \vspace{-0.5em}
    \label{fig:adjustp}
\end{figure}

We obtain positive examples for our dataset from two sources: VG(with positive examples in HA-DPO) and Sharegpt4V.
After extracting positive examples from ShareGPT4V, we found them to be too long.
To mitigate length bias, we used GPT4o-mini to rewrite them to match the length of the negative examples.
The prompt used is shown in \cref{fig:adjustp}.
For positive examples sourced from HA-DPO, after generating negative examples, we followed the original approach by rewriting the negative examples using GPT4o-mini.
The prompt used is shown in \cref{fig:rewritep}.
Also, we can adopt the method in HA-DPO to create more data.
For $k$ and $i$, we make an empirical choice based on performance and original settings.

\subsection{Long Context Hallucination}
\label{long-details}
\begin{figure}[h]
    \centering
    \includegraphics[width=\linewidth]{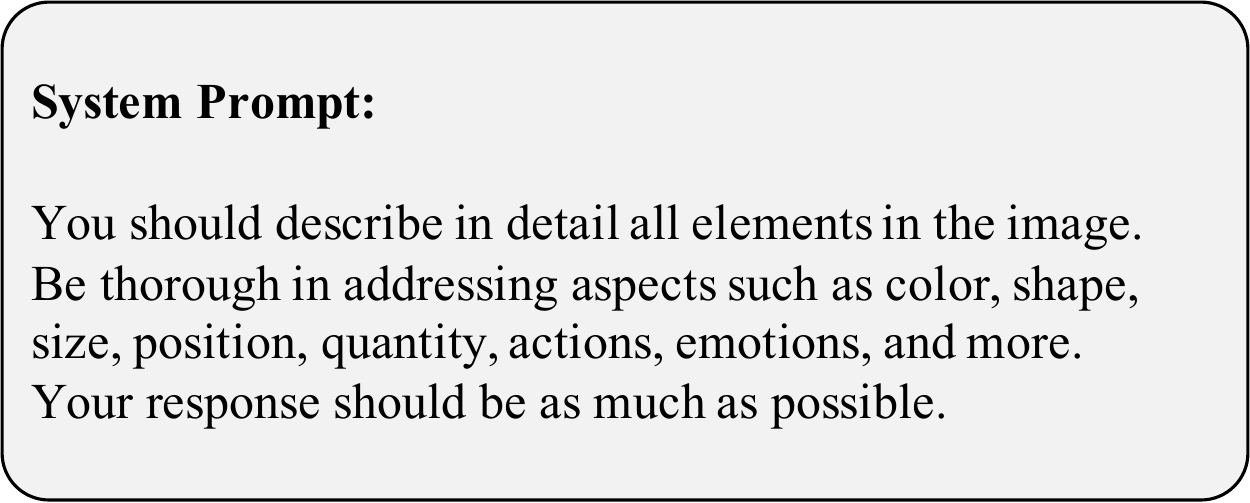}
    \vspace{-1.8em} %
    \caption{System Prompt used in \B}
    \vspace{-0.5em} %
    \label{fig:sysp}
\end{figure}

\begin{figure}[h]
    \centering
    \includegraphics[width=\linewidth]{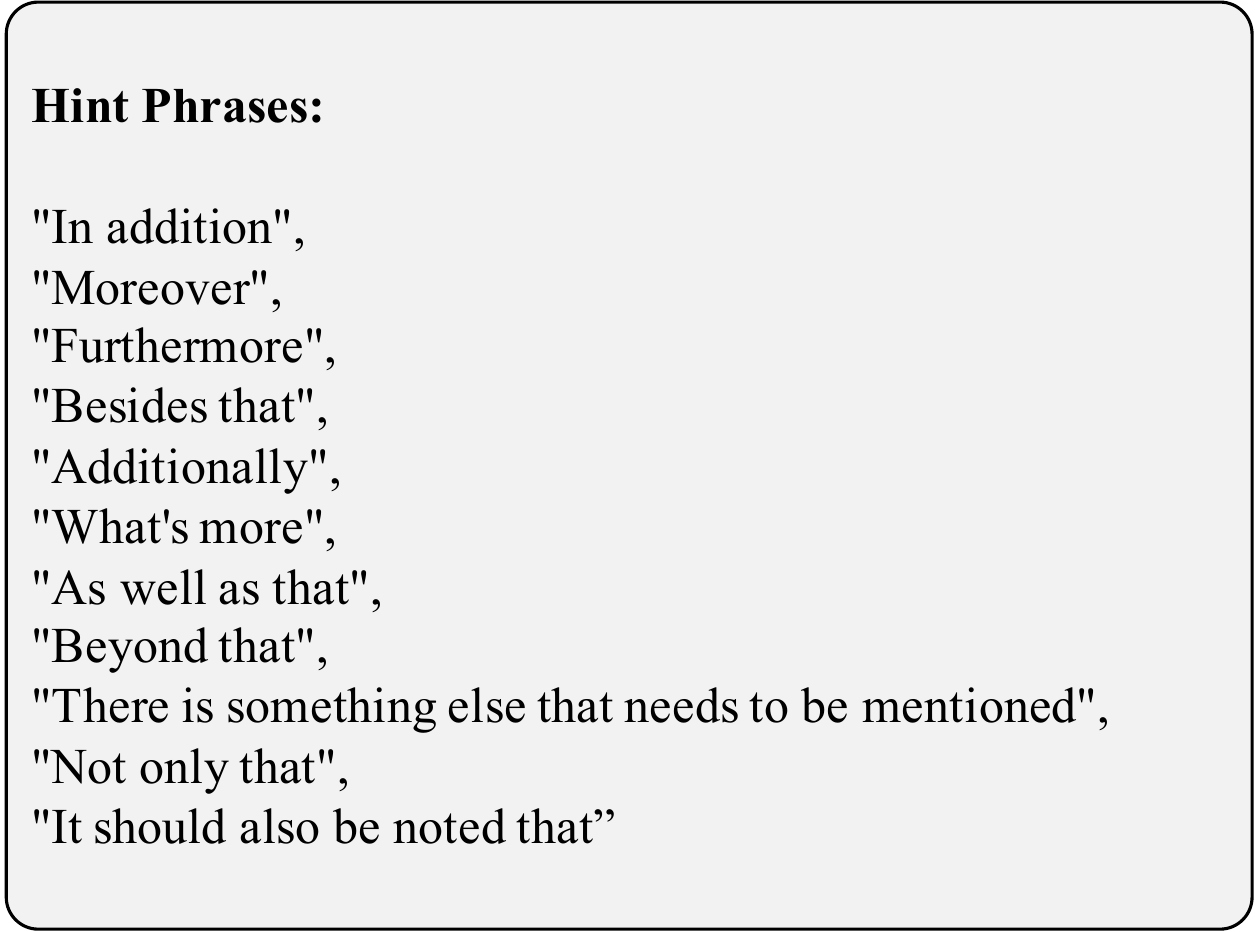}
    \vspace{-1.8em}
    \caption{Hint Phrases used in \B}
    \vspace{-0.5em}
    \label{fig:inducep}
\end{figure}

We used LLaVA-1.5-7B to continue generating text for the positive examples, with the system prompt in \cref{fig:sysp}, and the hint phrases in \cref{fig:inducep}.

By excluding the last two sentences, we aim to increase the concentration of hallucinated content in the tail of the response.
Generating three continuations at a time maintains an approximate balance in the average length between positive and negative examples.

\subsection{Multimodal Conflicts Hallucination}
\begin{figure}[h]
    \centering
    \includegraphics[width=\linewidth]{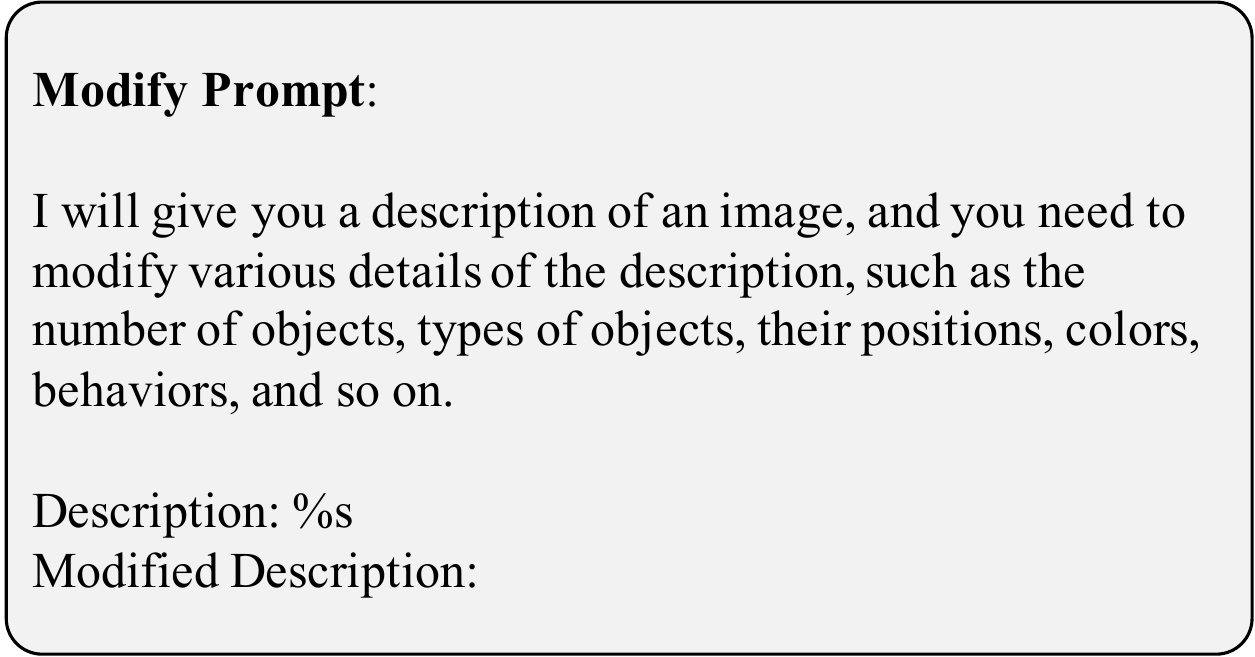}
    \vspace{-1.8em}
    \caption{Modify Prompt used in \C}
    \vspace{-0.5em}
    \label{fig:modifyp}
\end{figure}

We utilize GPT-4o-mini to modify the details of the positive examples, following the prompt shown in \cref{fig:modifyp}.
This approach introduces conflicting information that deviates from the image content.

\subsection{Effect of data ratio}
We did not conduct detailed experiments comparing different data type ratios.
However, throughout the experiments, all tested ratios showed significant improvements over the original model.
We report the best-performing dataset from our experiments.
Determining the optimal ratio of different data types is inherently a more challenging and general problem, which goes beyond the scope of this paper.

\FloatBarrier

\end{document}